\documentclass[final]{cvpr}

\usepackage{times}
\usepackage{epsfig}
\usepackage{graphicx}
\usepackage{amsmath}
\usepackage{amssymb}
\usepackage{enumerate}
\usepackage{color}
\usepackage{multirow}
\usepackage{amsthm,amsmath,amssymb}
\usepackage{mathrsfs}
\usepackage{amsopn}
\usepackage{amsmath,bm}
\usepackage{array}
\usepackage{booktabs}
\usepackage{algorithm}
\usepackage{subfig}
\usepackage{algpseudocode}
\usepackage{wrapfig}
\usepackage{times}
\usepackage{graphicx}
\usepackage{helvet}
\usepackage{courier}
\usepackage{wrapfig}
\usepackage{times}
\usepackage{graphicx}
\usepackage{helvet}

\newtheorem{myTheo}{Theorem}
\newcommand{\sign}{\text{sign}}

\usepackage[pagebackref=true,breaklinks=true,colorlinks,bookmarks=false]{hyperref}

\def\cvprPaperID{1896} 
\def\confYear{CVPR 2021}

\begin{document}

\title{ReNAS: Relativistic Evaluation of Neural Architecture Search}

\author{Yixing Xu$^1$, Yunhe Wang$^1$, Kai Han$^1$, Yehui Tang$^{14}$, Shangling Jui$^2$, Chunjing Xu$^1$, Chang Xu$^{3}$ \\
	$^1$Noah's Ark Lab, Huawei Technologies, $^2$Huawei Technologies\\
	$^3$The University of Sydney, $^4$Peking University\\
	\small{\texttt{\{yixing.xu, yunhe.wang\}@huawei.com;}} \small{\texttt{c.xu@sydney.edu.au}}
}

\maketitle
\pagestyle{empty}
\thispagestyle{empty}

\begin{abstract}
An effective and efficient architecture performance evaluation scheme is essential for the success of Neural Architecture Search (NAS). To save computational cost, most of existing NAS algorithms often train and evaluate intermediate neural architectures on a small proxy dataset with limited training epochs. But it is difficult to expect an accurate performance estimation of an architecture in such a coarse evaluation way. This paper advocates a new neural architecture evaluation scheme, which aims to determine which architecture would perform better instead of accurately predict the absolute architecture performance. Therefore, we propose a \textbf{relativistic} architecture performance predictor in NAS (ReNAS). We encode neural architectures into feature tensors, and further refining the representations with the predictor. The proposed relativistic performance predictor can be deployed in discrete searching methods to search for the desired architectures without additional evaluation. Experimental results on NAS-Bench-101 dataset suggests that, sampling 424 ($0.1\%$ of the entire search space) neural architectures and their corresponding validation performance is already enough for learning an accurate architecture performance predictor. The accuracies of our searched neural architectures on NAS-Bench-101 and NAS-Bench-201 datasets are higher than that of the state-of-the-art methods and show the priority of the proposed method.~\footnote{Published on CVPR 2021. \\    The pytorch code can be found at https://github.com/huawei-noah/Efficient-Computing/tree/master/Efficient-NAS/ReNAS}
\end{abstract}

\section{Introduction}
Recent years have witnessed the emergence of many well-known Convolutional Neural Networks (CNNs), (\eg VGG \cite{simonyan2014very}, ResNet \cite{he2016deep}, MobileNet \cite{howard2017mobilenets}). They have achieved state-of-the-art results in many real-world applications \cite{long2015fully,ren2015faster,simonyan2014very,zhou2018deep,tang2020scop,han2020ghostnet,chen2019data,chen2020learning}. However, the design of these sophisticated CNNs were heavily relied on human expert experience. Thus, it is attractive to investigate an automatic way to design neural network architectures without human intervention. Neural Architecture Search (NAS) has been proposed to address this need~\cite{cai2018proxylessnas,hu2020dsnas,lu2020nsganetv2,lu2021neural,yang2020searching,tang2020semi}.

Motivated by different searching strategies and assumptions, a number of NAS algorithms have been proposed to increase the search speed and the performance of the resulting network \cite{baker2017accelerating,hutter2011sequential,snoek2015scalable,li2016hyperband}, including discrete searching methods such as Evolutionary Algorithm (EA) based methods \cite{liu2017hierarchical,miikkulainen2019evolving,real2019regularized}, Reinforcement Learning (RL) based methods \cite{baker2016designing,pham2018efficient,zoph2016neural,zoph2018learning}, and continuous searching methods such as DARTS \cite{liu2018darts} and CARS \cite{yang2020cars}.

There have been a large body of works focusing on designing different searching methods. However, the architecture evaluation scheme has not been sufficiently studied yet. For the sake of evaluation efficiency, early stopping strategy is often adopted in the architecture evaluation phase of discrete searching methods. Based on the intermediate performance of a super-net, continuous searching methods optimize a series of learnable parameters to select layers or operations in deep neural networks. Nevertheless, these coarse architecture evaluations could prevent us from selecting the optimal neural architecture (detailed information can be found in Sec.3.1). A recent report suggested that the performances of the networks discovered by current NAS frameworks are similar to that of random search \cite{sciuto2019evaluating,yang2019evaluation}.

Most recently, there is an alternative way to evaluate neural architecture by learning a performance predictor. For example, Domhan \etal proposed a weight probabilistic model to extrapolate the performance from the first part of a learning curve \cite{domhan2015speeding}. Klein \etal used a Bayesian neural network to predict unobserved learning curves \cite{klein2016learning}. These two methods rely on Markov chain Monte Carlo (MCMC) sampling procedures and hand-crafted curve function, which are computationally expensive. Deng \etal~\cite{deng2017peephole} developed a unified way to encode individual layers into vectors and brought them together to form an integrated description via LSTM, and directly predicted the performance of a network architecture. Sun \etal~\cite{sun2019surrogate} proposed an end-to-end off-line performance predictor based on random forest.

\begin{figure*}[htb]
	\centering
	\includegraphics[width=0.99\linewidth]{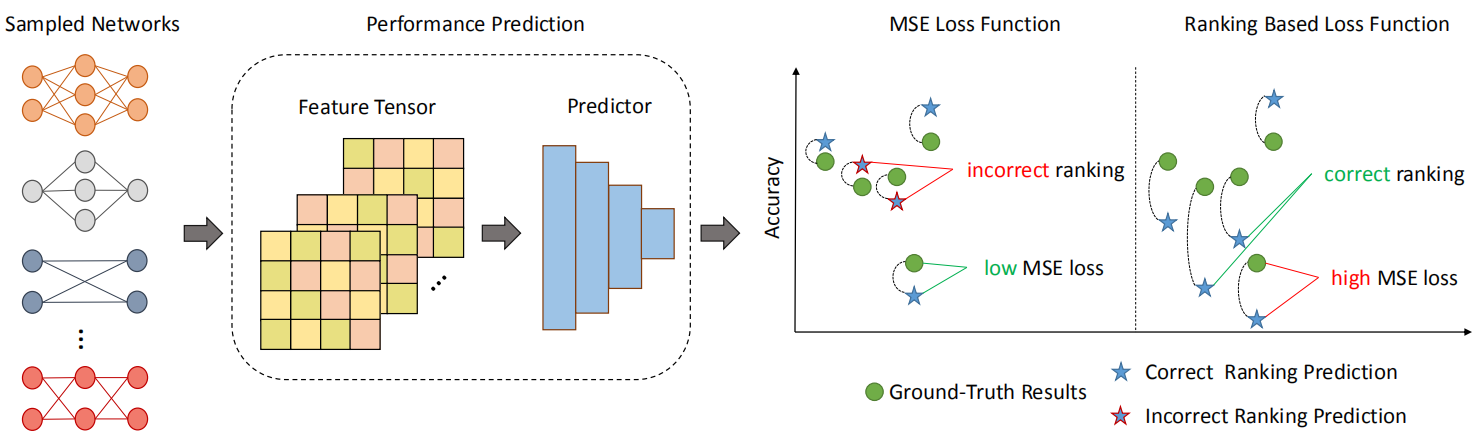}
	\caption{Pipeline of the proposed ReNAS. \textbf{Sampled Networks:} Architectures are sampled from the pre-defined search space and trained until converge to get the ground-truth results. \textbf{Performance Prediction.} Sampled architectures are encoded into feature tensors by leveraging the adjacency matrix and features that can well represent the computational power. Then, a predictor is used to predict the final accuracy. \textbf{Loss Function.} The pairwise ranking based loss function is used to train the predictor. Compared to mean squared error (MSE) loss, ranking based loss function may derive a prediction that is far from the ground-truth but in the correct ranking position, which is crucial to the following searching algorithms to search for the best architecture.}
	\label{figure_3}
\end{figure*}

Methods mentioned above focused on predicting the \emph{exact} performance of a given neural architecture with element-wise loss function such as mean squared error (MSE) or least absolute error (L1). However, in neural architecture search, what we care about is actually which neural architecture would lead to a better performance. Hence, instead of a deterministic evaluation of neural architectures, it is more reasonable to adopt a relativistic way to evaluate the performance of architectures.

In this paper, we aim to learn an architecture performance predictor that focuses on the rankings of different architectures. Specifically, given a cell-based search space with a unified super-net for all of the architectures (\eg NASBench \cite{ying2019bench,dong2020bench}, DARTS \cite{liu2018darts}), we produce a way to encode network architectures into tensors by leveraging the adjacency matrix of the cell and features that can well represent the computational power of a given architecture (\eg FLOPs and parameters of each node). Then, pairwise ranking based loss function is used instead of element-wise loss function, since keeping rankings between different networks are more important than accurately predicting their absolute performance for most of the searching methods. The pipeline of the proposed method is shown in Fig.~\ref{figure_3}. The experimental results on NAS-Bench-101 search space show that the proposed predictor achieves a higher prediction performance compared to the other state-of-the-art predictors, and can efficiently find the architecture with top $0.02\%$ accuracy in the whole search space by training only 424 ($0.1\%$ of the entire search space) neural architectures. Comparison with other state-of-the-art EA/RL/differential NAS methods on NAS-Bench-201 also shows the priority of the proposed method. Searched model on NAS-Bench-101 can be found in the MindSpore model zoo~\footnote{https://www.mindspore.cn/resources/hub/}.

\section{Problem Formulation}
In this section, we first instantiate the problem of evaluation scheme used in the previous methods. Then, we give an elaborate introduction of the proposed performance predictor. Specifically, developing architecture performance predictor consists of three parts: encoding network architecture into a feature tensor, the regressor (predictor) to predict the performance and the objective function to be optimized. In this paper, we propose a new approach to encode neural network architectures in cell-based search space into feature tensors and to design the regressor. Furthermore, we propose the pair-wise ranking loss to optimize the regressor.

\subsection{Evaluation Scheme}
In case of saving computational resource and time, a commonly used evaluation scheme in the previous NAS methods is to train the neural network $N$ on part of the training dataset $\tilde{D} \in D$ with early stop strategy, in which $D$ is the entire training dataset. The model is then tested on the validation set, and the intermediate accuracy $ACC(N, \tilde{D})$ stands for an approximation of the performance of the model in the subsequent searching algorithms instead of the ground-truth accuracy $ACC(N, {D})$. Previous methods~\cite{li2020random,yu2019evaluating,yang2019evaluation} assumed that there is a linear relationship between the intermediate accuracy and the ground-truth accuracy:
\begin{equation}
ACC(N, {D}) = k\times ACC(N, \tilde{D}) + \sigma,
\label{assumption}
\end{equation}
where $k$ is the scaling factor and $\sigma$ is the offset. However, the assumption in Eq. \ref{assumption} may not hold in practice, and the intermediate accuracy may break the original rankings between architectures since lighter architectures often converge faster on smaller dataset than cumbersome architectures, but perform worse when using the whole training set \cite{sciuto2019evaluating}. Note that producing correct rankings for the searching algorithm is rather important, since the searching algorithms always select relatively better architectures regardless of their absolute performance.

Thus, we focus on learning the correct rankings between different architectures with predictor. Specifically, given a predictor $\varepsilon$ and two different architectures $N_1$ and $N_2$. Denote $\varepsilon(N; \mathcal W)$ as the predicted performance of a given architecture $N$ in which $\mathcal W$ is the weight matrix of the predictor, we should have:
\begin{equation}
\begin{aligned}
\varepsilon(N_1; \mathcal W)&>\varepsilon(N_2; \mathcal W), \\
\text{if and only if} \ \ ACC(&N_1, {D})>ACC(N_2, {D}),
\label{correct}
\end{aligned}
\end{equation}
which means that the predictor should rank different network architectures into the right order according to their ground-truth performance.
\subsection{Feature Tensor of Cell-Based Search Space}

Commonly, a cell-based search space with a unified super-net stacks the same searched cell to get the final architecture \cite{liu2018darts,ying2019bench,dong2020bench}. In this section, we give an introduction on encoding the architectures in cell-based search space into feature tensors.

Encoding a neural architecture is important for a predictor to predict the performance. Peephole \cite{deng2017peephole} chose layer type, kernel width, kernel height and channel number as the representations of each layer. E2EPP \cite{sun2019surrogate} forced the network architecture to be composed of the DenseNet blocks, ResNet blocks and pooling blocks, and generated features based on these blocks.

However, those features are not strong enough to encode a network architecture. Different from the methods mentioned above, we focus on encoding architectures into feature tensors by leveraging the adjacency matrix of the cell and features that can well represent the computational power of a given architecture. In the following, we use NAS-Bench-101 dataset \cite{ying2019bench} as an example, which contains over 423k unique CNN architectures and their train, validation and test accuracies on CIFAR-10 dataset. The same methods can be applied on other cell-based search space.

Different cells produce different CNN architectures. In each cell there are no more than 7 nodes in which $IN$ and $OUT$ nodes are fixed to represent the input and output tensors of the cell, respectively. The other nodes are randomly selected from 3 different operations: $3\times3$ convolution, $1\times1$ convolution and $3\times3$ max-pooling. The edges are limited to no more than 9. Specifically, the cell can be represented by a 0-1 adjacency matrix $\mathcal A\in \{0,1\}^{n\times n}$ and a type vector ${\bf t}\in\{1,\cdot\cdot\cdot,5\}^n$ (5 different node types containing input, $3\times3$ conv, $1\times1$ conv, $3\times3$ max-pooling and output), in which $n$ is the number of nodes. Furthermore, we calculate FLOPs and parameters of each node and derive a FLOP vector ${\bf f}\in\mathbb R^n$ (we assume the input image size is $32\times32$) and a parameter vector ${\bf p}\in\mathbb R^n$.

Since the number of nodes may be different in each cell, we pad the adjacency matrix $\mathcal A$ with 0 and the size is fixed as $7\times7$. The type vector $\bf t$, FLOP vector $\bf f$ and parameter vector $\bf p$ are padded accordingly. Note that the input and the output node should be fixed as the first and last node, thus the zero-padding is added at penultimate row and column each time until the size of $\mathcal A$ is $7\times7$. After that, we broadcast the vectors into matrices, and make an element-wise multiplication with the adjacency matrix to get the type matrix $\bf T$, FLOP matrix $\bf F$ and parameter matrix $\bf P$, and finally concatenate them together to get a $19\times7\times7$ tensor $\mathcal T$ to represent a specific architecture in NAS-Bench-101. An example of the process of deriving feature tensor is shown in Sec.2 of the supplementary material.

Note that the feature tensor representations are not robust to permutation, \ie, permuting the adjacency and type matrices may lead to different results. This problem can be solved by fixing the order of the nodes. Specifically, we sort nodes by distances to INPUT node with depth-first-search in order to reduce the non-unique ordering phenomenon. For those nodes with the same depth, we investigate a simple data augmentation method (\ie, permuting the adjacency and type matrices of the same architecture based on the nodes with the same depth) so that all of the representations for a specific architecture are assigned with the same label. 

\subsection{Architecture Performance Predictor}
Given the feature tensor mentioned above, we propose the architecture performance predictor and introduce the ranking based loss function.

In practice, there are usually limited training data for the predictor due to the massive time and resource spent on training a single neural architecture. Thus, in order to prevent the over-fitting problem, we use a simple LeNet-5 architecture to predict the final accuracy of a given network architecture tensor $\mathcal T$.

When training the predictor, a commonly used loss function is element-wise MSE or L1 loss function \cite{deng2017peephole,istrate2018tapas,sun2019surrogate}. They assume that a lower MSE or L1 loss leads to a better ranking results. However, this is not always the case. For example, given two networks with ground-truth classification accuracies of 0.9 and 0.91 on the validation set. In the first circumstance, they are predicted to have accuracies of 0.91 and 0.9, and in the second circumstance of 0.89 and 0.92. MSE losses are the same in both situations, but the former is worse since the ranking between two networks is changed and the architecture with worse performance will be selected by the searching methods. We believe that the rankings of predicted accuracies between different architectures are more important than their absolute performance when applying network performance predictor to different searching methods.

Formally, given $n$ different network architectures and their ground-truth performance $\{(N_i,y_i)\}_{i=1}^n$, and $\{\varepsilon(N_i; \mathcal W)\}_{i=1}^n$ is the output of the predictor (short as $\varepsilon(N_i)$), which are the $n$ objects to be ranked. We define the pairwise ranking based loss function as:
\begin{equation}
{\mathcal L_1}(\mathcal W)=\sum_{i=1}^{n-1} \sum_{j=i+1}^{n} {\phi((\varepsilon(N_i)-\varepsilon(N_j))*sign(y_i-y_j))},
\label{fcn_1}
\end{equation}
in which $\phi(z) = (a-z)_+$ is hinge function with parameter $a$. Given a pair of examples, the loss is 0 only when the examples are in the correct order and differed by a margin. Other functions like logistic function $\phi(z)=log(1+e^{-z})$ and exponential function $\phi(z)=e^{-z}$ can also be applied here.

Besides utilizing the final output, we believe that the feature extracted before the last FC layer is also useful. The continuity is a common assumption in machine learning, \ie the performance changes continuously along the feature space. However, this is not the case for primary network architecture, in which a slightly change of the architecture may lead to a radical change of the performance (\eg skip connect). Thus, we consider learning the feature with the property of continuity.

In order to generate features with continuity, consider the triplet $\{(\eta(N_i;\mathcal W), y_i)\}_{i=1}^3$ in which $\eta(N_i;\mathcal W)$ (short as $\eta(N_i)$) is the feature generated before the final FC layer. The Euclidean distance between the two features is computed as $d_{ij} = ||\eta(N_i)-\eta(N_j)||_2$, and the difference of the performance between two architectures is simply computed as $l_{ij}=|y_i-y_j|$. Thus, we achieve the property of continuity by defining the loss function as:
\begin{equation}
{\mathcal L_2}(\mathcal W)=\sum_{i=1}^{n-2} \sum_{j=i+1}^{n-1} \sum_{k=j+1}^{n} {\phi((d_{ij}-d_{ik})*sign(l_{ij}-l_{ik}))}.
\label{fcn_2}
\end{equation}
Given a single triplet, there are several different pairs, and the pair with smaller distance (smaller $d_{ij}$) should have similar performance (smaller $l_{ij}$). Eq.(4) compares two different pairs, and produces a cost when the former pair has larger distance (bigger $d_{ij}$) but similar performance (smaller $l_{ij}$) compared to the latter, and vice versa. The loss is accumulated on all different triplets.

Note that although the form of Eq. \ref{fcn_1} and Eq. \ref{fcn_2} are similar, the purposes behind are quite different. Given the equations above, the final loss function is the combination of them:
\begin{equation}
{\mathcal L} = \mathcal L_1 + \lambda\mathcal L_2,
\label{fcn_3}
\end{equation}
in which $\lambda$ is the hyper-parameter that controls the importance between two different loss functions. Therefore, the effects of the proposed predictor are two folds. The first is to directly predict accuracies with correct ranking, the second is to generate features with the property of continuity which indirectly helps to predict the accuracies.

Finally, the performance predictor is integrated into discrete searching algorithms such as EA (RL) based searching method by replacing the fitness (reward) of a given architecture with the output of our predictor (see Fig.~\ref{figure_3}). EA based method is used in the following experiments, and the individual is fed into the predictor and the output is treated as the fitness of the model in EA method within milliseconds.

\section{Theoretical Analysis}
In this section, we analyze the generalization error bound and prove that using the proposed pairwise ranking based loss function (Eq.~\ref{fcn_1}) is better than using MSE loss when solving the ranking problem, under the assumption of using a two layer neural network with ReLU activation function.

Firstly, we reformulate the ranking based loss function. Given an input pair $(x,y)$, $(x',y')\in(\mathcal X\times\mathcal Y)^2$, denote $f:\mathcal X\rightarrow\mathbb R$ as the ranking function on $\mathcal X$, and $\ell:\mathbb R\times(\mathcal X\times\mathcal Y)^2\rightarrow \mathbb \{0,R^+\}$ be the ranking loss function, the expected error of $f$ can be defined as~\cite{agarwal2009generalization}:
\begin{equation}
R_\ell(f) = \mathbb E_{((X,Y),(X',Y'))\sim(\mathcal X\times\mathcal Y)^2}[\ell(f,(X,Y),(X',Y'))].
\end{equation}
Given a training set $\mathcal D=\{x_i,y_i\}_{i=1}^n\in\{\mathcal X,\mathcal Y\}^n$, the empirical error of $f$ is defined as:
\begin{equation}
\hat R_\ell(f) = \frac{1}{n(n-1)}\sum_{i=1}^{n-1}\sum_{j=i+1}^{n}\ell(f,(x_i,y_i),(x'_i,y'_i)),
\end{equation}
and the regularized empirical error is defined as:
\begin{equation}
\hat R_\ell^\lambda(f) = \hat R_\ell(f) + \lambda C(f),
\end{equation}
in which the second term is the regularization term and $\lambda>0$ is the regularization parameter.

Thus, when using the loss function $\phi(z)=(a-z)_+$, Eq.~\ref{fcn_1} equals to using a hinge ranking loss which is denoted as:
\begin{equation}
\ell_{{h}}(f,(x,y),(x',y'))=[a-(f(x)-f(x'))\cdot \sign(y-y')]_+,
\end{equation}
and the element-wise MSE loss can be denote as:
\begin{equation}
\ell_{\text{mse}}(f,(x,y),(x',y'))=\frac{1}{2}[(f(x)-y)^2+(f(x')-y')^2].
\end{equation}

In the following, we give the generalization error bounds when using pairwise ranking based loss function and MSE loss, and show that the proposed loss function is better. The proof is applied in the supplementary material.

\begin{myTheo}
	Given $\mathcal A$ as the symmetric ranking algorithm\footnote{The output of a symmetric ranking algorithm is independent of the order of elements in the training sequence $\mathcal D$. The proposed algorithm can be easily proved to be a symmetric ranking algorithm.} whose outputs of samples on a training dataset $\mathcal D\in(\mathcal X\times\mathcal Y)^n$ is $f_{\mathcal D}=\arg\min_{f\in\mathcal F}\hat R_\ell^\lambda(f)$, in which $n\in\mathbb N$ is the number of training samples. Denote $c_x$ and $c_f$ as the upper bound of the inputs and weights such that for all $x\in\mathcal X$ and $f:\mathcal X\rightarrow \mathbb R$ we have $|x|\leq c_x$ and $\|f\|_2\leq c_f$. Also given $\ell_h$ as the hinge ranking loss function that satisfy $0\leq \ell_h(f,(x,y),(x',y'))\leq L$ for all $f:\mathcal X\rightarrow \mathbb R$ and $(x,y),(x',y')\in(\mathcal X\times\mathcal Y)^2$, and $\ell_{\text{mse}}$ as the MSE loss function that also satisfy $0\leq \ell_{\text{mse}}(f,(x,y),(x',y'))\leq L$. Then for any $0<\delta<1$, with probability at least $1-\delta$ we have:
	\begin{align}
	R_{\ell_h}(f_\mathcal D)&<\hat R_{\ell_h}(f_\mathcal D)+\frac{8c_x^2c_f^2}{\lambda n}  \notag\\
	&+ (\frac{4c_x^2c_f^2}{\lambda}+L) \sqrt{\frac{2\ln(1/\delta)}{n}},
	\end{align}
	and
	\begin{align}
	R_{\ell_{\text{mse}}}(f_\mathcal D)&<\hat R_{\ell_\text{mse}}(f_\mathcal D)+\frac{8(\frac{c_xc_fL}{2\sqrt\lambda}+1)c_x^2c_f^2}{\lambda n} \notag\\
	&+ (\frac{4(\frac{c_xc_fL}{2\sqrt\lambda}+1)c_x^2c_f^2}{\lambda}+L) \sqrt{\frac{2\ln(1/\delta)}{n}}.
	\end{align}
\end{myTheo}
Since $(\frac{c_xc_fL}{2\sqrt\lambda}+1)>1$, also note that the lower the gap between the expected error and the empirical error, the better the generalization ability. Thus, we can say that using pairwise ranking based loss function (Eq.~\ref{fcn_1}) has a better generalization ability than using element-wise MSE loss.

\section{Experiments}
In this section, we conduct several experiments on verifying the effectiveness of the proposed network performance predictor. After that, the best CNN architecture is found by embedding the predictor into EA algorithm and is compared to other state-of-the-art predictors to verify its performance.

The parameter settings for training the predictor and searching for best architecture are detailed below. When training the predictor, we used Adam to train the LeNet architecture with initial learning rate of $1\times10^{-3}$; the weight decay is set to $5\times10^{-4}$; the batch size is set to $1024$ and trained for $200$ epochs. When using the EA algorithm, we set the maximum generation number to $500$ and population size to $64$. The probability for selection, crossover and mutation are set to $0.5$, $0.3$ and $0.2$, respectively.

\subsection{Predictor Performance Comparison on NAS-Bench-101}
We compared the proposed predictor with the methods introduced in Peephole \cite{deng2017peephole} and E2EPP \cite{sun2019surrogate}. The NAS-Bench-101 dataset is selected as the training and testing sets of the predictors. 

Recall that one of the fundamental idea in ReNAS is that the ranking of the predicted values is more important than their absolute values when embedding the predictor into different searching methods. Thus, for the quantitative comparison, we use the Kendall's Tau (KTau) \cite{sen1968estimates} as the indicator:
\begin{equation}
\text{KTau} = 2\times\frac{\text{number of concordant pairs}}{C_n^2}-1,
\end{equation}
in which $n$ is the number of samples, $C_n^2=n(n-1)/2$ and concordant pair means the rankings of predicted values and the actual values of a given pair are the same. KTau ranges from $-1$ to $1$ and is suitable for judging the quality of the predictive rankings. A higher value indicates a better ranking.

In order to clearly review the influence of using feature tensor and pairwise loss, we conduct the following $6$ versions of the proposed methods by fixing the predictor as LeNet and varying the feature encoding method and the loss function, including:

\textbf{ReNAS-1 (type matrix + MSE)}: Using only the type matrix as feature and MSE loss function.

\textbf{ReNAS-2 (tensor + MSE)}: Using the proposed feature tensor and MSE loss function.

\textbf{ReNAS-3 (type matrix + $\mathcal L_1$)}: Using only the type matrix as feature and loss function $\mathcal L_1$ (Eq. \ref{fcn_1}).

\textbf{ReNAS-4 (tensor + $\mathcal L_1$)}: Using the proposed feature tensor and loss function $\mathcal L_1$.

\textbf{ReNAS-5 (type matrix + $\mathcal L$)}: Using only the type matrix as feature and loss function $\mathcal L$ (Eq. \ref{fcn_3}).

\textbf{ReNAS-6 (tensor + $\mathcal L$)}: Using the proposed feature tensor and loss function $\mathcal L$.

\begin{table*}[t]
	\renewcommand\arraystretch{1.05}
	\caption{The Kendall's Tau (KTau) of Peephole, E2EPP and the proposed algorithms on the NAS-Bench-101 dataset with different proportions of training samples.}
	\label{table_3}
	\centering
	\begin{tabular}{l|c|c|c|c|c|c|c}
		\hline
		Methods & $0.1\%$ & $1\%$ & $10\%$ & $30\%$  & $50\%$ & $70\%$ & $90\%$\\
		\hline\hline
		Peephole \cite{deng2017peephole}& 0.4556 & 0.4769 & 0.4963 & 0.4977 & 0.4972 & 0.4975 & 0.4951 \\
		E2EPP \cite{sun2019surrogate}& 0.5038 & 0.6734 & 0.7009 & 0.6997 & 0.7011 & 0.6992 & 0.6997 \\
		\hline
		ReNAS-1 & 0.3465 & 0.5911 & 0.7914 & 0.8229& 0.8277& 0.8344& 0.8350\\
		ReNAS-2 & 0.4856 & 0.6090 & 0.8103 & 0.8430 & 0.8399 & 0.8504& 0.8431 \\
		ReNAS-3 & 0.6039 & 0.7943 & 0.8752 & 0.8894& 0.8949& 0.8976& 0.8995\\
		ReNAS-4 & 0.6335 & 0.8136 & 0.8762 & \textbf{0.8900} & \textbf{0.8957} & \textbf{0.8979} & \textbf{0.8997} \\
		ReNAS-5 & 0.6096 & 0.7949 & 0.8756 &0.8854 &0.8898 &0.8911 & 0.8918\\
		ReNAS-6 & \textbf{0.6574} & \textbf{0.8161} & \textbf{0.8763} & 0.8873 & 0.8910 & 0.8923 & 0.8954 \\
		\hline
	\end{tabular}
\end{table*}

Note that the search space in NAS-Bench-101 dataset and E2EPP are different from each other, and the encoding method proposed in E2EPP is unable to be used directly on NAS-Bench-101 dataset. In order to apply NAS-Bench-101 dataset to E2EPP, we produce surrogate method for E2EPP and use the feature encoding method proposed in the previous section instead of the original encoding method produced by E2EPP. The other parts remain unchanged.

The experimental results are shown in Tab. \ref{table_3}. Consider that the training samples can only cover a small proportion of the search space in reality, we focus on the second column when using only $0.1\%$ (424 models and the corresponding validation accuracies) of the NAS-Bench-101 dataset as training set. Different proportions are used in the experiment for integrity. The results show that the proposed encoding method can represent an architecture better than without using it, and the KTau indicator increases about $0.14$ when using MSE loss and $0.05$ when using pairwise loss. When using pairwise loss instead of element-wise MSE loss, the KTau indicator increases about $0.26$ when using only the type matrix as feature, and about $0.17$ when using the proposed feature tensor. It means that pairwise loss is better than MSE loss at ranking regardless of input feature.

Comparing to other state-of-the-art methods, Peephole used kernel size and channel number as features in addition to layer (node) type, and shows better result than ReNAS-1 method which uses only the layer (node) type as features. However, it performs worse than ReNAS-2 method when using all the feature proposed, which again shows the superiority of using feature tensors. E2EPP used random forest as predictor, which has advantages only when the training samples are extremely rare. When using limited training data, the proposed method with loss function $\mathcal L$ (Eq. \ref{fcn_3}) achieves the best KTau performance, while the proposed method with $\mathcal L_1$ loss (Eq. \ref{fcn_1}) is better when more training data is used. The results show that generating features with continuity has advantageous to model ranking when little training data is used, which is often the case in reality.

\begin{figure*}[t]
	\subfloat[Peephole]{
		\begin{minipage}[t]{0.33\linewidth}
			\centering
			\includegraphics[width=0.9\linewidth]{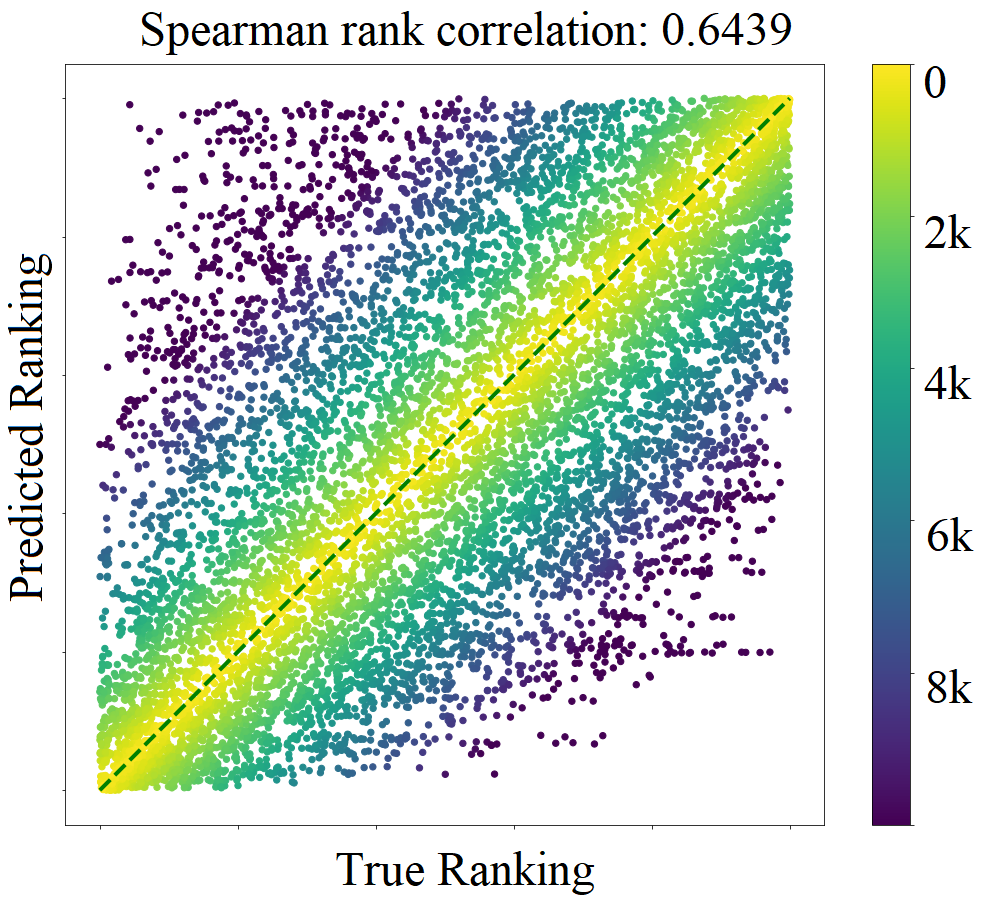}
		\end{minipage}
	}
	\subfloat[E2EPP]{
		\begin{minipage}[t]{0.33\linewidth}
			\centering
			\includegraphics[width=0.9\linewidth]{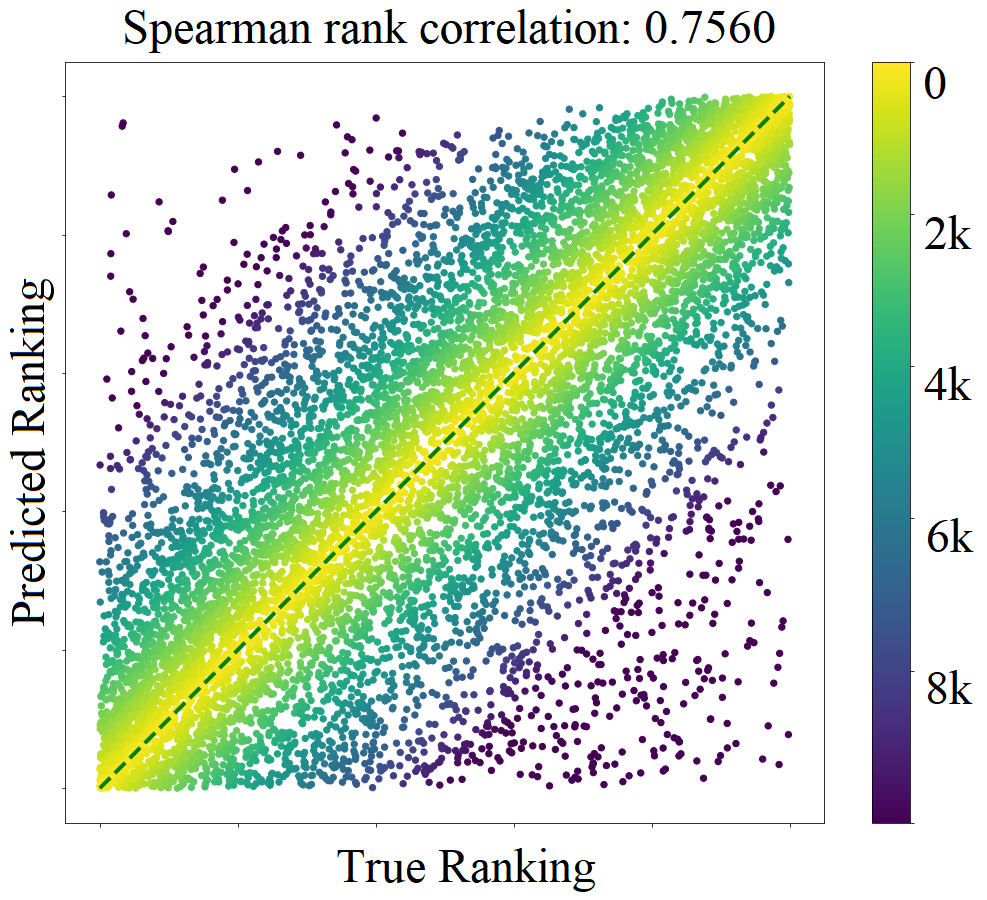}
		\end{minipage}
	}
	\subfloat[Proposed]{
		\begin{minipage}[t]{0.33\linewidth}
			\centering
			\includegraphics[width=0.9\linewidth]{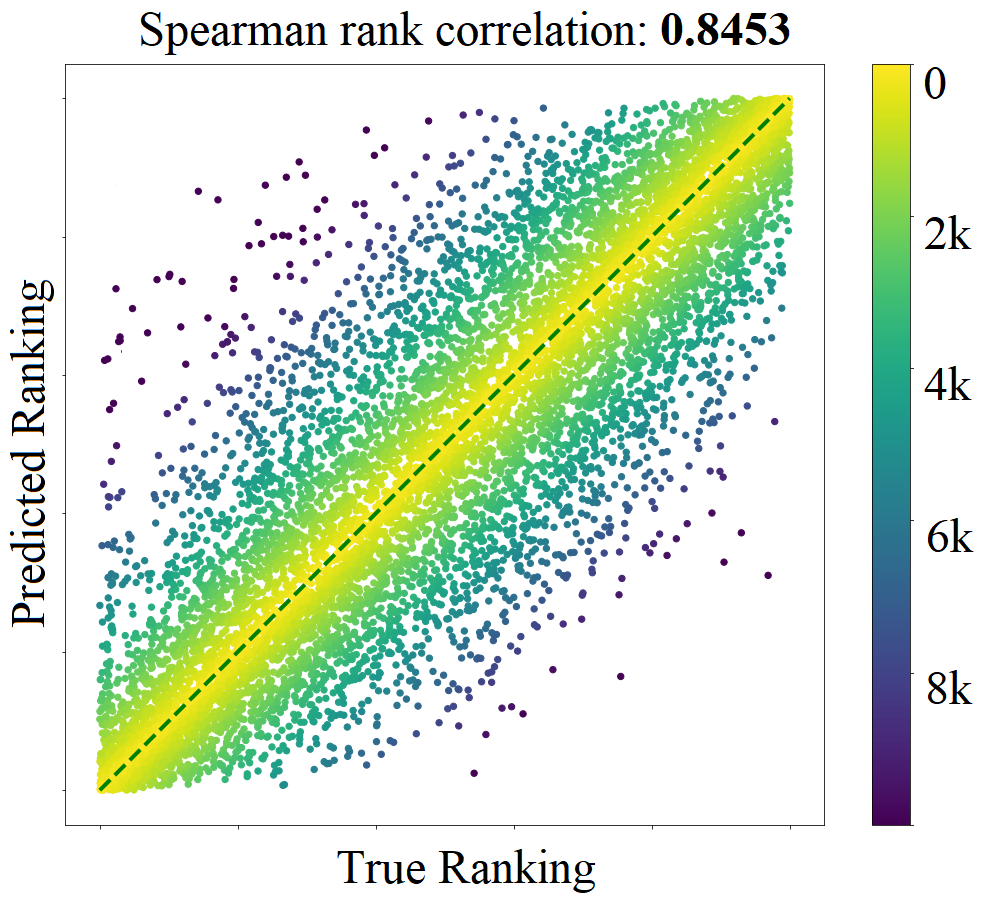}
		\end{minipage}
	}
	\caption{The predicted ranking and true ranking of Peephole, E2EPP and the proposed method on NAS-Bench-101 dataset. 1000 models are randomly selected for exhibition purpose. The $x$ axis denotes the true ranking, and the $y$ axis denotes the corresponding predicted ranking.}
	\label{figure_4}
\end{figure*}

A qualitative comparisons on NAS-Bench-101 dataset is shown in Fig.~\ref{figure_4}. We show the results of training predictors using $0.1\%$ training data, the $x$ axis of each point represents the true ranking among all the points and the $y$ axis denotes the corresponding predicted ranking. The points made by a perfect predictor lie on the line $y=x$, and the closer the points to line $y=x$ the better. The results show that the predicted ranking made by ReNAS is better than other state-of-the-art methods.
\begin{table}[t]
		\centering
		\renewcommand\arraystretch{1.05}
		\caption{The classification accuracy ($\%$) on CIFAR-10 dataset and the ranking ($\%$) among different architectures in NAS-Bench-101 dataset using EA algorithm with the proposed predictor and the peer competitors. Predictors are trained with $0.1\%$ samples randomly selected from NAS-Bench-101 dataset.}
		\label{table_4}
		\vspace{-0.0em}
		\begin{tabular}{c|c|c}
			\hline
			Method & accuracy($\%$) & ranking($\%$)\\
			\hline\hline
			Peephole~\cite{deng2017peephole} & 92.63 $\pm$ 0.31& 12.32\\
			E2EPP~\cite{sun2019surrogate} & 93.47 $\pm$ 0.44& 1.23\\
			RS & 93.72 $\pm$ 0.13 & 0.23\\
			\hline
			ReNAS-1 & 92.36 $\pm$ 0.27 & 16.93\\
			ReNAS-2 & 93.03 $\pm$ 0.21& 6.09\\
			ReNAS-3 & 93.43 $\pm$ 0.26& 1.50\\
			ReNAS-4 & 93.90 $\pm$ 0.21 & 0.04\\
			ReNAS-5 & 93.48 $\pm$ 0.18 & 1.21\\
			ReNAS-6 & \textbf{93.95 $\pm$ 0.11} & \textbf{0.02}\\
			\hline
		\end{tabular}
\end{table}

\subsection{Architecture Search Results on NAS-Bench-101}
When searching for the best architecture, the size of the training set of the predictor should be limited since the search space in EA algorithm is the same as in NAS-Bench-101, and we cannot prevent EA algorithm from searching the architectures in the training set. Thus, in order to reduce the influence of the training set, we used only $0.1\%$ of NAS-Bench-101 dataset as training samples to train the predictor, and subsequently used for EA algorithm. The final performance tested on CIFAR-10 dataset with the best architecture searched by EA algorithm with the proposed predictor, the results of random search (RS) and the peer competitors mentioned above are shown in Tab. \ref{table_4}.
Specifically, the best performances among top-10 architectures selected by EA algorithm with different predictors are reported and the experiments are repeated 20 times with different random seed to alleviate the randomness.

\begin{table}[t]
	\renewcommand\arraystretch{1.05}
	\caption{The classification accuracy ($\%$) on CIFAR-10 dataset and the ranking ($\%$) among different architectures in NAS-Bench-101 dataset using the predictors trained with $0.1\%$ samples selected from NAS-Bench-101 dataset. Different selection methods are used.}
	\label{table_5}
	\centering
	\begin{tabular}{c|c|c}
		\hline
		Method & accuracy($\%$) & ranking($\%$) \\
		\hline\hline
		random selection & \textbf{93.95 $\pm$ 0.11} & \textbf{0.02} \\
		select by parameters & 93.84 $\pm$ 0.21& 0.08\\
		select by FLOPs & 93.76 $\pm$ 0.13 & 0.16\\
		\hline
	\end{tabular}
\end{table}

The second column represents the classification accuracies of the selected models on CIFAR-10 test set, and the third column represents the true ranking of the selected models among all the $423k$ different models in NAS-Bench-101 dataset. The proposed method outperforms other competitors, and finds an network architecture with top $0.02\%$ performance among the search space using only $0.1\%$ dataset. The fact of achieving good performance with little training data is reasonable for two reasons. The first is that the fundamental features of FLOPs and parameters can represent the architecture well and tensor like input is suitable for CNN. The second is that using pairwise loss expand the training set to some extent. Given $n$ individuals, there are actually $n(n-1)/2$ pairs and $n(n-1)(n-2)/6$ triplets for training.

\begin{table}[t]
	\centering
	\renewcommand\arraystretch{1.05}
	\caption{The classification accuracy ($\%$) on CIFAR-100 dataset among different architectures in NAS-Bench-101 using EA algorithm with the proposed predictor and the peer competitors. Predictors are trained with 424 samples randomly selected from NAS-Bench-101.}
	\label{table_8}
	\vspace{-0.0em}
	\begin{tabular}{c|c|c}
		\hline
		Method & top-1 acc (\%)  & top-5 acc (\%) \\
		\hline\hline
		Peephole~\cite{deng2017peephole} & 73.58& 91.97\\
		E2EPP~\cite{sun2019surrogate}  &75.49 & 92.77\\
		RS &77.47 & 93.68\\
		\hline
		Proposed &\textbf{78.56} & \textbf{94.17}\\
		\hline
	\end{tabular}
\end{table}

Note that when using performance predictor in practice, the search space is often different from NAS-Bench-101 dataset, which means the training samples needs to be collected from scratch. Thus, we give some intuitions of selecting model architectures from search space as training samples. $0.1\%$ samples are selected from NAS-Bench-101 dataset as training samples with the method of random selection, select by parameters and select by FLOPs. When selecting by parameters (FLOPs), all samples are sorted by their total parameters (FLOPs), and selected uniformly. Different predictors are trained with different training samples using proposed method, and are further integrated into EA algorithm for searching. The performance of the best architectures are shown in Tab. \ref{table_5}.
The results show that random selection performs best. A possible reason is that architectures with similar parameters (FLOPs) perform diversely, and the uniformly selected architectures cannot represent the true performance distribution of the architectures with similar parameters (FLOPs). Thus, random selection is our choice, and is worth trying when generating training samples from search space in reality.

\begin{table*}[t]
	\renewcommand\arraystretch{1.05}
	\caption{Search results on NAS-Bench-201.}
	\label{table_201}
	\centering
	\begin{tabular}{c|c|c|c|c|c|c|c}
		\hline
		\multirow{2}{*}{Method} & Search & \multicolumn{2}{c|}{CIFAR-10} & \multicolumn{2}{c|}{CIFAR-100} &  \multicolumn{2}{c}{ImageNet-16-120}\\
		\cline{3-8}
		& seconds & validation & test& validation & test& validation & test\\
		\hline
		RSPS~\cite{li2019random} & 7587.12 & 84.16$\pm$1.69 & 87.66$\pm$1.69 & 59.00$\pm$4.60& 58.33$\pm$4.34& 31.56$\pm$3.28& 31.14$\pm$3.88\\
		DARTS-V1~\cite{liu2018darts} & 10889.87 & 39.77$\pm$0.00 & 54.30$\pm$0.00 & 15.03$\pm$0.00 & 15.61$\pm$0.00 & 16.43$\pm$0.00 & 16.32$\pm$0.00\\
		DARTS-V2~\cite{liu2018darts} & 29901.67 & 39.77$\pm$0.00 & 54.30$\pm$0.00 & 15.03$\pm$0.00 & 15.61$\pm$0.00 & 16.43$\pm$0.00 & 16.32$\pm$0.00\\
		GDAS~\cite{dong2019searching} & 28925.91 & 90.00$\pm$0.21 & 93.51$\pm$0.13 & 71.15$\pm$0.27 & 70.61$\pm$0.26 & 41.70$\pm$1.26 & 41.84$\pm$0.90\\
		SETN~\cite{dong2019one} & 31009.81 & 82.25$\pm$5.17 & 86.19$\pm$4.63 & 56.86$\pm$7.59 & 56.87$\pm$7.77 & 32.54$\pm$3.63 & 31.90$\pm$4.07\\
		ENAS~\cite{pham2018efficient} & 13314.51 & 39.77$\pm$0.00 & 54.30$\pm$0.00 & 15.03$\pm$0.00 & 15.61$\pm$0.00 & 16.43$\pm$0.00 & 16.32$\pm$0.00 \\
		\hline
		NPENAS~\cite{wei2020npenas} & - & 91.08$\pm$0.11 & 91.52$\pm$0.16 & -&-&-&-\\
		REA~\cite{real2019regularized} & 0.02 & \textbf{91.19$\pm$0.31} & 93.92$\pm$0.30 & 71.81$\pm$1.12 & 71.84$\pm$0.99 & 45.15$\pm$0.89 & 45.54$\pm$1.03\\
		RS & 0.01 & 90.03$\pm$0.36 & 93.70$\pm$0.36 & 70.93$\pm$1.09 & 71.04$\pm$1.07 & 44.45$\pm$1.10 & 44.57$\pm$1.25\\
		NASBOT~\cite{white2020study} & - & - & 93.64$\pm$0.23 & - & 71.38$\pm$0.82 & - & 45.88$\pm$0.37\\
		REINFORCE~\cite{williams1992simple} & 0.12& 91.09$\pm$0.37 & 93.85$\pm$0.37 & 71.61$\pm$1.12 & 71.71$\pm$1.09 & 45.05$\pm$1.02 & 45.24$\pm$1.18\\
		BOHB~\cite{falkner2018bohb} & 3.59 & 90.82$\pm$0.53 & 93.61$\pm$0.52 & 70.74$\pm$1.29 & 70.85$\pm$1.28 & 44.26$\pm$1.36 & 44.42$\pm$1.49\\
		ReNAS(ours) & 86.31& 90.90$\pm$0.31 & \textbf{93.99$\pm$0.25} & \textbf{71.96$\pm$0.99} & \textbf{72.12$\pm$0.79} & \textbf{45.85$\pm$0.47} & \textbf{45.97$\pm$0.49}\\
		\hline
		ResNet & \multirow{2}{*}{N/A} & 90.83& 93.97& 70.42& 70.86& 44.53& 43.63\\
		\cline{1-1}\cline{3-8}
		optimal & & 91.61& 94.37& 73.49& 73.51& 46.77& 47.31\\
		\hline
	\end{tabular}
\end{table*}

We further extend our method to search space with unknown labels by searching for architecture in NAS-Bench-101 search space that performs well on CIFAR-100 dataset. Specifically, we randomly select 424 architectures and train them on CIFAR-100 from scratch and get the ground-truth labels. These samples are further used to train the predictor, and the best architecture is searched using the methods mentioned above. The results in Tab. \ref{table_8} show the priority of the proposed method. Other experiments on NAS-Bench-101 dataset are given in the supplementary material.

\subsection{Compare with other NAS Search methods on NAS-Bench-201}
In order to compare with other state-of-the-art NAS search methods, we further conduct experiments on NAS-Bench-201~\cite{dong2020bench}, which is also a cell-based search space including 15625 different architectures and corresponding train, validation and test accuracies on CIFAR-10, CIFAR-100 and ImageNet-16-120~\cite{chrabaszcz2017downsampled} datasets. During the experiment, $90$ randomly selected architectures and the corresponding validation accuracies are used as the training set for the proposed predictor. After the predictor is trained, we traverse the search space with the predictor instead of using EA algorithm, since the number of architectures is small. Other settings are the same as the experiments on NAS-Bench-101 dataset. The best validation and test accuracies among top-10 architectures selected by the predictor are reported. Experiments are repeated 20 times with different training samples selected.

The comparative methods include: (1) Random Search methods, such as random search (RS) and random search with parameter sharing (RSPS)~\cite{li2019random}. (2) EA methods, such as REA~\cite{real2019regularized} and NPENAS~\cite{wei2020npenas}. (3) RL methods, such as REINFORCE~\cite{williams1992simple} and ENAS~\cite{pham2018efficient}. (4) Differentiable methods, such as DARTS-V1/DARTS-V2~\cite{liu2018darts}, GDAS~\cite{dong2019searching} and SETN~\cite{dong2019one}. (5) HPO methods, such as BOHB~\cite{falkner2018bohb}. (6) Predictor methods, such as NASBOT~\cite{white2020study}. Experimental settings of the comparative methods are the same as in~\cite{dong2020bench}, and the search results on validation sets and test sets for each dataset are shown in Tab.~\ref{table_201}.

The search cost of ReNAS is the training time of the predictor. Traversing the search space with the predictor is finished within milliseconds and is negligible compared to the training time. The search results show that the proposed ReNAS method produces state-of-the-art searching accuracy on all three datasets based on the test set with an acceptable search cost within two minutes on a single GeForce GTX 1080 Ti GPU, which indicate the superiority of the proposed method. Compared to the previous state-of-the-art method REA~\cite{real2019regularized}, NASBOT~\cite{white2020study} and random search, ReNAS finds better architectures that is 0.07\%, 0.35\% and 0.29\% better on CIFAR-10 test set, 0.28\%, 0.74\% and 1.08\% better on CIFAR-100 test set, and 0.43\%, 0.09\% and 1.40\% better on ImageNet-16-120 test set.

\section{Conclusion}
We proposed a new method for predicting network performance based on its architecture before training. We encode an architecture in cell-based search space into a feature tensor by leveraging the adjacency matrix of the cell and features that can well represent the computational power of a given architecture. The pairwise ranking based loss function is used for the performance predictor instead of the element-wise loss function, since the rankings between different architectures are more important than their absolute values in different searching methods. We also theoretically proved the superiority of using pairwise ranking loss. Several experiments are conducted on NAS-Bench-101 dataset, and shows the priority of the proposed predictor on sorting the performance of different architectures and searching for an architecture with top performance among the search space using only $0.1\%$ of the dataset. Experimental results on NAS-Bench-201 dataset shows that the proposed ReNAS outperform state-of-the-art NAS searching methods with a considerable search cost.

\subsubsection*{Acknowledgment}
We thank anonymous area chair and reviewers for their helpful comments. Chang Xu was supported by the Australian Research Council under Project DE180101438.

{\small
\bibliographystyle{ieee_fullname}
\bibliography{ref}
}

\end{document}


\title{Supplementary Material\\ReNAS: Relativistic Evaluation of Neural Architecture Search}

\author{Yixing Xu$^1$, Yunhe Wang$^1$, Kai Han$^1$, Yehui Tang$^{14}$, Shangling Jui$^2$, Chunjing Xu$^1$, Chang Xu$^{3}$ \\
	$^1$Noah's Ark Lab, Huawei Technologies, $^2$Huawei Technologies\\
	$^3$The University of Sydney, $^4$Peking University\\
	\small{\texttt{\{yixing.xu, yunhe.wang\}@huawei.com;}} \small{\texttt{c.xu@sydney.edu.au}}
}

\maketitle

\section{Proof of Theorem 1}
We first give the definition of $\sigma$-admissibility of the ranking loss function $\ell$:
\newtheorem{defnision}{Definition}
\begin{defnision}
	\textbf{($\sigma$-admissibility)} Given $\mathcal F$ as a class of real-valued functions on $\mathcal X$. Denote $\ell$ as the ranking loss function and $\sigma>0$. Then $\ell$ is $\sigma$-admissible with respect to $\mathcal F$, if for all $f_1, f_2\in \mathcal F$ and all $(x,y), (x',y')\in(\mathcal X\times\mathcal Y)$, we have:
	\begin{align}
	|\ell(f_1,(x,y),(x',y'))-\ell(f_2,(x,y),(x',y'))|\leq\notag\\ \sigma(|f_1(x)-f_2(x)|+|f_1(x')-f_2(x')|).
	\end{align}
\end{defnision}

Then, we will get the following generalization error bound for a given ranking loss function $\ell$:
\newtheorem{lemma}{Lemma}
\begin{lemma}
	Given $\mathcal A$ as the symmetric ranking algorithm whose outputs of samples on a training dataset $\mathcal D\in(\mathcal X\times\mathcal Y)^n$ is $f_{\mathcal D}=\arg\min_{f\in\mathcal F}\hat R_\ell^\lambda(f)$, in which $n\in\mathbb N$ is the number of training samples. Denote $c_x$ and $c_f$ as the upper bound of the inputs and weights such that for all $x\in\mathcal X$ and $f:\mathcal X\rightarrow \mathbb R$ we have $|x|\leq c_x$ and $\|f\|_2\leq c_f$. Also given $\ell$ as the ranking loss function that satisfy $0\leq \ell(f,(x,y),(x',y'))\leq L$ for all $f:\mathcal X\rightarrow \mathbb R$ and $(x,y),(x',y')\in(\mathcal X\times\mathcal Y)^2$. Then for any $0<\delta<1$, with probability at least $1-\delta$ we have:
	\begin{equation}
	R_{\ell}(f_\mathcal D)<\hat R_{\ell}(f_\mathcal D)+\frac{8\sigma c_x^2c_f^2}{\lambda n} + (\frac{4\sigma c_x^2c_f^2}{\lambda}+L) \sqrt{\frac{2\ln(1/\delta)}{n}}.
	\end{equation}
\end{lemma}
\begin{proof}
	Given the assumption of using a two layer neural network with ReLU activation function, we can denote the output of the neural network as:
	\begin{equation}
	f(x)=W_2\eta(W_1\cdot x),
	\end{equation}
	in which $W_1$ and $W_2$ are the parameters of the given network, and $\eta$ indicates the ReLU activation function. Also denote $\|f\|_2=\sqrt{\|W_1\|_2^2+\|W_2\|_2^2}$ as the $\ell_2$-norm of the parameters, we then have:
	\begin{align}
	|f(x)|&=|W_2\eta(W_1\cdot x)|\notag\\
	&\leq \big||W_2|\eta(|W_1\cdot x|) \big|\notag\\
	&=|W_2\cdot W_1\cdot x|\notag\\
	&=|W_2\cdot W_1|\cdot |x|\notag\\
	&\leq \frac{1}{2}(\|W_1\|_2^2+\|W_2\|_2^2)|x|\notag\\
	&\leq \frac{1}{2}c_xc_f\|f\|_2.
	\label{ineq1}
	\end{align}
	Thus, given Fcn.~\ref{ineq1} mentioned above, and Theorem.8, Fcn.6 and Theorem.11 in~\cite{agarwal2009generalization}, we can successfully prove Lemma 1.
\end{proof}

After that, we prove that a hinge ranking loss is $1$-admissible with respect to $\mathcal F$ and an MSE loss is $(\frac{c_xc_f L}{2\sqrt\lambda}+1)$-admissible with respect to $\mathcal F$.

\newtheorem{theorem}{Theorem}
\begin{myTheo}
	Given $\mathcal F$ as a class of real-valued functions on $\mathcal X$. Denote $\ell$ as the ranking loss function and $\sigma>0$. Then $\ell_h(f,(x,y),(x',y'))=[(a-(f(x)-f(x'))\cdot \sign(y-y'))]_+$ is $1$-admissible with respect to $\mathcal F$, \eg for all $f_1, f_2\in \mathcal F$ and all $(x,y), (x',y')\in(\mathcal X\times\mathcal Y)$, we have:
	\begin{align}
	|\ell_h(f_1,(x,y),(x',y'))-\ell_h(f_2,(x,y),(x',y'))|\leq\notag\\ |f_1(x)-f_2(x)|+|f_1(x')-f_2(x')|.
	\end{align}
\end{myTheo}

\begin{proof}
	Without loss of generality, we assume that $\ell_h(f_1,(x,y),(x',y'))\geq\ell_h(f_2,(x,y),(x',y'))$. Note that when $\ell_h(f_1,(x,y),(x',y'))=\ell_h(f_2,(x,y),(x',y'))$, we simply have:
	\begin{align}
	|\ell_h(f_1,(x,y),(x',y'))-\ell_h(f_2,(x,y),(x',y'))|=0\leq\notag\\|f_1(x)-f_2(x)|+|f_1(x')-f_2(x')|,
	\end{align} 
	thus the following prove is based on $\ell_h(f_1,(x,y),(x',y'))>\ell_h(f_2,(x,y),(x',y'))$, and can be divided into following situations:
	
	(1) $(f_1(x)-f_1(x'))\cdot \sign(y-y')\leq a$ and $(f_2(x)-f_2(x'))\cdot \sign(y-y')\leq a$. Then we have:
	\begin{align}
	&|\ell_h(f_1,(x,y),(x',y'))-\ell_h(f_2,(x,y),(x',y'))|\notag\\
	=& |a-(f_1(x)-f_1(x'))\cdot \sign(y-y')\notag\\
	&-a+(f_2(x)-f_2(x'))\cdot \sign(y-y')|\notag\\
	=& \sign(y-y')|f_1(x)-f_2(x)+f_1(x')-f_2(x')|\notag\\
	\leq& |f_1(x)-f_2(x)+f_1(x')-f_2(x')|\notag\\
	\leq& |f_1(x)-f_2(x)|+|f_1(x')-f_2(x')|.
	\end{align}
	(2) $(f_1(x)-f_1(x'))\cdot \sign(y-y')\leq a$ and $(f_2(x)-f_2(x'))\cdot \sign(y-y')> a$. Then we have:
	\begin{align}
	&|\ell_h(f_1,(x,y),(x',y'))-\ell_h(f_2,(x,y),(x',y'))|\notag\\
	=& |a-(f_1(x)-f_1(x'))\cdot \sign(y-y')-0|\notag\\
	<& |a-(f_1(x)-f_1(x'))\cdot \sign(y-y')\notag\\
	&-(a-(f_2(x)-f_2(x'))\cdot \sign(y-y'))|\notag\\
	\leq& |f_1(x)-f_2(x)|+|f_1(x')-f_2(x')|.
	\end{align}
	Therefore, in all situations we have:
	\begin{align}
	|\ell_h(f_1,(x,y),(x',y'))-\ell_h(f_2,(x,y),(x',y'))|\leq\notag\\
	 |f_1(x)-f_2(x)|+|f_1(x')-f_2(x')|,
	\end{align}
	and thus $\ell_h(f_1,(x,y),(x',y'))$ is $1$-admissible with respect to $\mathcal F$.
\end{proof}

\begin{myTheo}
	Given $\mathcal F$ as a class of real-valued functions on $\mathcal X$. Denote $\ell$ as the ranking loss function and $\sigma>0$. Then $\ell_{\text{mse}}(f,(x,y),(x',y'))=\frac{1}{2}((f(x)-y)^2+(f(x')-y')^2)$ is $(\frac{c_xc_f L}{2\sqrt\lambda}+1)$-admissible with respect to $\mathcal F$, \eg for all $f_1, f_2\in \mathcal F$ and all $(x,y), (x',y')\in(\mathcal X\times\mathcal Y)$, we have:
	\begin{align}
	|\ell_h(f_1,(x,y),(x',y'))-\ell_h(f_2,(x,y),(x',y'))|\leq\notag\\
	 (\frac{c_xc_f L}{2\sqrt\lambda}+1)\big(|f_1(x)-f_2(x)|+|f_1(x')-f_2(x')|\big).
	\end{align}
\end{myTheo}

\begin{proof}
	\begin{align}
	&|\ell_{\text{mse}}(f_1,(x,y),(x',y'))-\ell_{\text{mse}}(f_2,(x,y),(x',y'))|\notag\\
	=& \frac{1}{2}|(f_1(x)-y)^2+(f_1(x')-y')^2\notag\\
	&+(f_2(x)-y)^2+(f_2(x')-y')^2| \notag\\
	=& \frac{1}{2}|f_1^2(x)-2f_1(x)+f_1^2(x')-2f_1(x')\notag\\
	&-f_2^2(x)+2f_2(x)-f_2^2(x')+2f_2(x')|\notag\\
	=& \frac{1}{2}|f_1^2(x)-f_2^2(x)+f_1^2(x')-f_2^2(x')\notag\\
	&-2(f_1(x)-f_2(x))-2(f_1(x')-f_2(x'))|\notag\\
	\leq& \frac{1}{2}(|f_1^2(x)-f_2^2(x)|+|f_1^2(x')-f_2^2(x')|\notag\\
	&+2|(f_1(x)-f_2(x))|+2|(f_1(x')-f_2(x'))|)\notag\\
	=& \frac{1}{2}(|f_1(x)+f_2(x)|+2)|f_1(x)-f_2(x)|+\notag\\
	&\frac{1}{2}(|f_1(x')+f_2(x')|+2)|f_1(x')-f_2(x')|.
	\label{eq0}
	\end{align}
	Given Fcn.~\ref{ineq1} mentioned above, we have:
	\begin{equation}
	|f(x)|\leq \frac{1}{2}c_xc_f\|f\|_2.
	\label{eq1}
	\end{equation}
	
	Also note that:
	\begin{align}
	\ell_{\text{mse}}=\hat R_{\ell_\text{mse}} + \lambda \|f\|_2^2 \leq L.
	\end{align}
	Since $\hat R_{\ell_\text{mse}}>0$, we have:
	\begin{align}
	\|f\|_2^2 \leq \frac{L^2}{\lambda}.
	\label{eq2}
	\end{align}
	Applying Eq.~\ref{eq2} to Eq.~\ref{eq1}, we have:
	\begin{equation}
	|f(x)|\leq \frac{1}{2}c_xc_f\|f\|_2 \leq \frac{c_xc_f L}{2\sqrt\lambda}.
	\label{eq3}
	\end{equation}
	Finally, applying Eq.~\ref{eq3} to Eq.~\ref{eq0}, we can derive the $(\frac{c_xc_f L}{2\sqrt\lambda}+1)$-admissibility of $\ell_{\text{mse}}$:
	\begin{align}
	&|\ell_{\text{mse}}(f_1,(x,y),(x',y'))-\ell_{\text{mse}}(f_2,(x,y),(x',y'))|\notag\\
	\leq& \frac{1}{2}(|f_1(x)+f_2(x)|+2)|f_1(x)-f_2(x)|\notag\\
	&+\frac{1}{2}(|f_1(x')+f_2(x')|+2)|f_1(x')-f_2(x')|\notag\\
	\leq& (\frac{c_xc_f L}{2\sqrt\lambda}+1)\big(|f_1(x)-f_2(x)|+|f_1(x')-f_2(x')|\big),
	\end{align} 
	and thus finish the proof.
\end{proof}
Combining the above definition, lemma and theorems, we have proved Theorem 1 in the main paper.

\section{An Example of Deriving Feature Tensor}
\begin{figure*}[htb]
	\centering
	\includegraphics[width=0.99\linewidth]{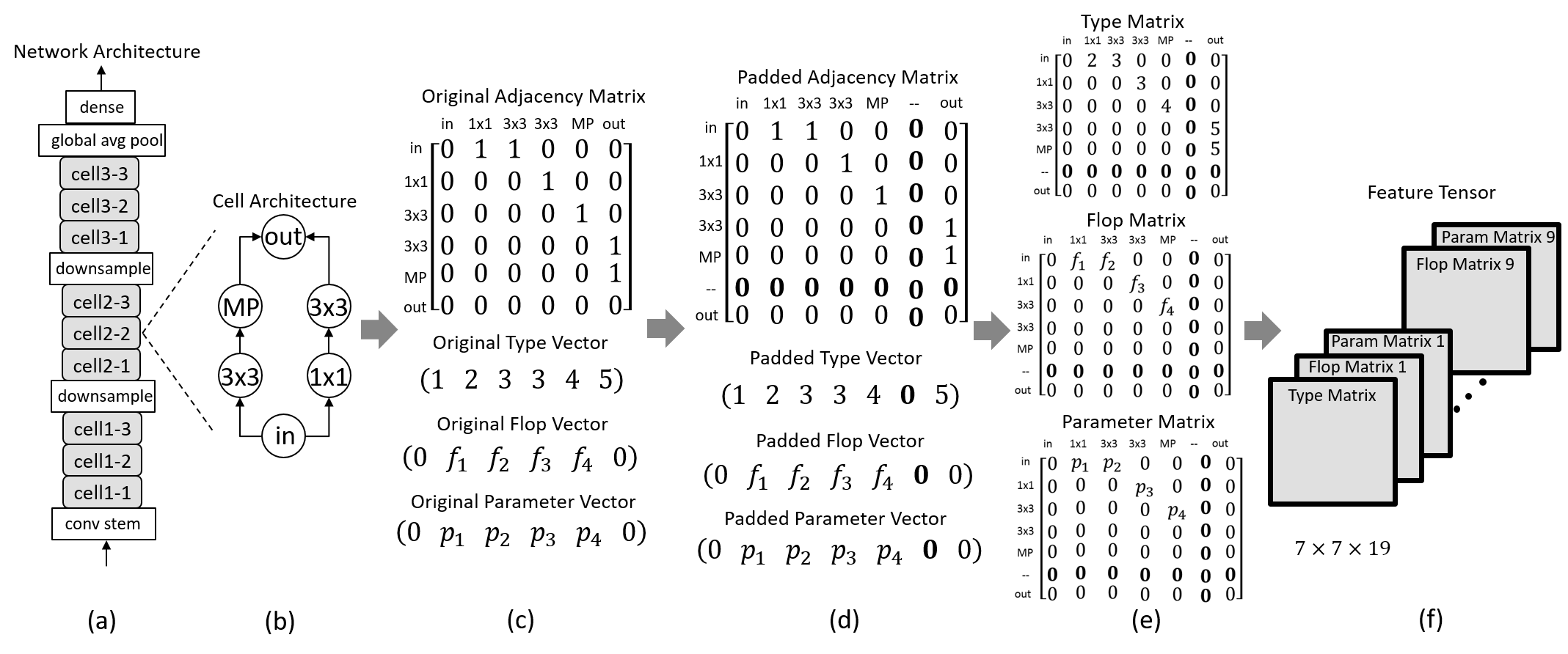}
	\caption{An example of encoding neural network architecture into feature tensor. \textbf{(a)}: The skeleton of the neural network architecture. \textbf{(b)}: A specific cell architecture with 6 nodes. \textbf{(c)}: The corresponding adjacency matrix $\mathcal A$, type vector $\bf t$, FLOP vector $\bf f$ and parameter vector $\bf p$ of the cell. \textbf{(d)}: Padding adjacency matrix $\mathcal A$ to $7\times7$ and vectors accordingly. Note that the zero-padding is added at penultimate row and column, since the last row and column represents the output node. \textbf{(e)}: Vectors are broadcasted into matrix, and an element wise multiplication is made with the adjacency matrix to get the type matrix, FLOP matrix and parameter matrix. \textbf{(f)}: There are 9 cells in the network, thus producing 9 different FLOP matrices and parameter matrices. All the cells share the same type matrix. We concatenate all the matrices to get the final $19\times7\times7$ tensor.}
	\label{figure_3}
\end{figure*}

In this section, we give an example of the process of deriving feature tensor from cell-based search space NAS-Bench-101. We use an architecture with $6$ nodes in a cell, and the process is shown in Fig.~\ref{figure_3}.

\begin{figure*}[t]
	\centering
	\setlength{\abovecaptionskip}{-0.0cm}
	\setlength{\belowcaptionskip}{0.0cm}
	\includegraphics[width=1\linewidth]{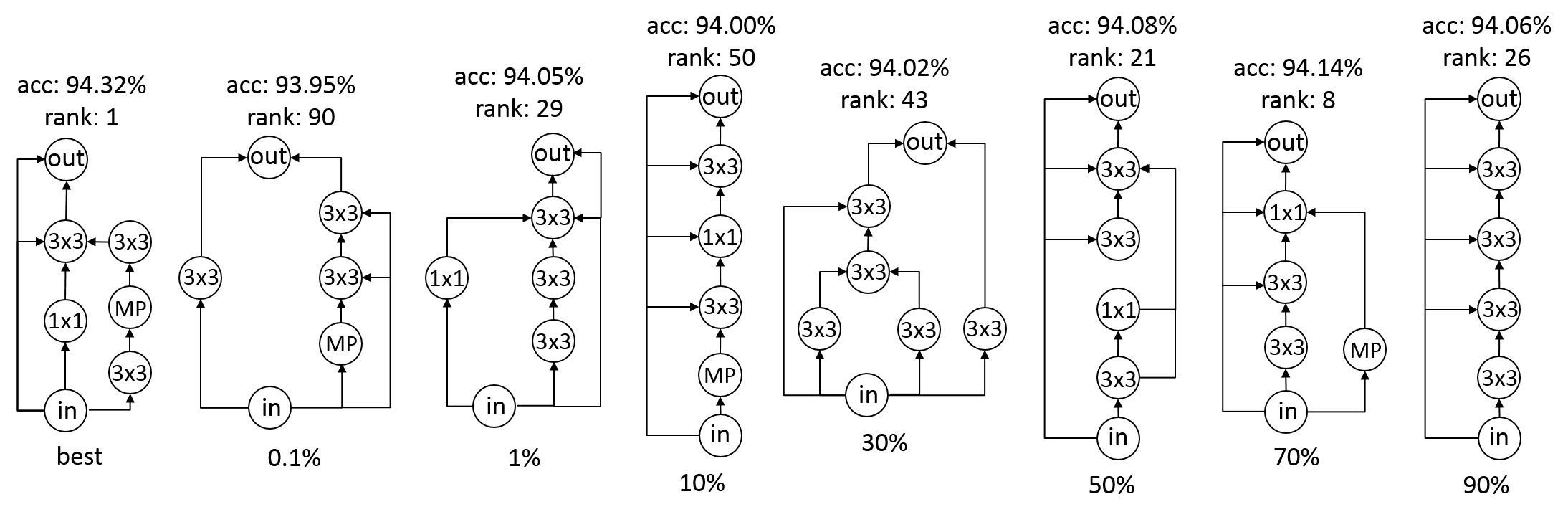}
	\caption{The best architectures found by the predictor with different ratio of training samples.}
	\label{figure_6}
\end{figure*}
\section{More Experiments on NAS-Bench-101}
In this section, we conduct more experiments on NAS-Bench-101 dataset to further demonstrate the usefulness of the proposed ReNAS method.

In the following we give an intuitive representation of the best architectures selected by the performance predictor with different number of training samples as shown in Fig. \ref{figure_6}. The best cell architecture searched by EA algorithm using proposed predictor trained with random selected training samples is shown in column 2 of Fig. \ref{figure_6}. Note that the rank-1 architecture in NAS-Bench-101 dataset cannot be selected by the predictor even when using $90\%$ of the training data. This is because when using pairwise ranking based loss function, there are $n(n-1)/2$ training pairs and it is inefficient to train them in a single batch. Thus, mini-batch updating method is used and a single architecture is compared with limited architectures in one epoch, which causes the lack of global information about this architecture especially when the number of training samples is large. In fact, the mini-batch size $b$ is set to 1024 in the experiment, and it is a compromise between effectiveness and efficiency.

\begin{table}[t]
	\renewcommand\arraystretch{1.2}
	\caption{Statistics on NAS-Bench-101 dataset. `$3\times3$' refers to whether the model uses this operation. `Distance' refers to the distance between input node and output node. `$\#$model' refers to the number of models. `Best acc' refers to the performance of the best architecture among `$\#$model' number of models on CIFAR-10 dataset. `Average acc' refers to the average performance of `$\#$model' number of models on CIFAR-10 dataset.}
	\label{table_6}
	\centering
	\begin{tabular}{c|c|c|c|c}
		\hline
		$3\times3$ & Distance & $\#$model & Best acc & Average acc\\
		\hline\hline
		\multirow{6}{*}{yes}    & 1 &   68552   &   \textbf{94.32}   &   \textbf{91.97}   \\
		& 2 &   153056  &   94.05   &   91.02   \\
		& 3 &   110863  &   93.68   &   89.31   \\
		& 4 &   27227   &   92.36   &   87.40   \\
		& 5 &   2516    &   90.54   &   86.51   \\
		& 6 &   211     &   88.87   &   84.91   \\
		\hline
		\multirow{6}{*}{no}     & 1 &   12468   &   91.62   &   88.40   \\
		& 2 &   26282   &   90.81   &   86.69   \\
		& 3 &   17735   &   90.24   &   83.53   \\
		& 4 &   4282    &   88.95   &   80.20   \\
		& 5 &   400     &   88.16   &   78.84   \\
		& 6 &   32      &   86.71   &   74.93   \\
		\hline
	\end{tabular}
\end{table}

\begin{table}[t]
	\renewcommand\arraystretch{1.2}
	\caption{Predictors trained and evaluated with the whole NAS-Bench-101 dataset and sub dataset. The experiments are repeated 20 times to alleviate the randomness of the results.}
	\label{table_7}
	\centering
	\begin{tabular}{c|c|c}
		\hline
		Datasets &  accuracy($\%$) & ranking($\%$) \\
		\hline\hline
		whole dataset & 93.95 $\pm$ 0.11& 0.02\\
		sub dataset & \textbf{94.02 $\pm$ 0.14}& \textbf{0.01}\\
		\hline
	\end{tabular}
\end{table}

This is the same reason that the performance of the architecture found by the predictor trained with $90\%$ dataset is marginally better than that trained with $0.1\%$ dataset. Specifically, we divide the architectures into two parts. The first part is the architectures trained with $0.1\%$ and $1\%$ dataset, and the second part is the rest. Note that in the first part the number of training sample is on the same order of magnitude with the mini-batch size $b$, thus the global information of a single model is easy to obtain and the performance becomes better when there are more training data. In the second part, the number of training sample is significantly larger than $b$. On one hand, increasing the number of samples helps training. On the other hand, the global ranking information is harder to get. Thus, the performance is marginally better when using more training samples.

Finally, there are some common characteristics among these architectures. The first is that the distance between input and output node is at most 2, which shows the significance of skip-connection. The second is that $3\times3$ operation appears in each architecture. Based on these observations, we separate the NAS-Bench-101 dataset based on the distance between input node and output node, and whether the $3\times3$ operation is used. Some statistics are shown in Tab. \ref{table_6}.

It shows that the shorter the distance between input node and output node, the better the performance is. Besides, $3\times3$ operation helps the architecture to perform better. Based on the observation above, we may form a better search space of NAS-Bench-101 dataset by using only 68552 models with $3\times3$ operation and skip-connect between input node and output node. An experiment of training and evaluating performance predictor is conducted on this sub search space and the results show that the predictor trained and evaluated within the sub search space performs better than the previous one as shown in Tab. \ref{table_7}. It shows that a better search space helps to produce a better performance predictor.

{\small
	\bibliographystyle{ieee_fullname}
	\bibliography{ref}
}